%% file: main.tex
\documentclass[sigconf]{acmart}

\usepackage{hyperref}
\usepackage{url}
\usepackage{graphicx}
\usepackage{subcaption}
\usepackage{booktabs}
\usepackage{array}
\usepackage{multirow}
\usepackage{makecell}
\usepackage{colortbl}
\usepackage[T1]{fontenc}
\usepackage{tabularx}
\usepackage{enumitem}

\newcommand{\hide}[1]{}

\newcommand{\todo}[1]{{\color{red}{[[[ #1 ]]]}}}
\newcommand{\old}[1]{{\color{brown}{[OLD: #1]}}}

\newcommand{\oldinput}[1]{{\color{brown} #1} \\ \\}
\newcommand{\li}[1]{{\color{red}{\bf lc: #1}}}
\newcommand{\polo}[1]{{\color{red}{\bf pc: #1}}}
\newcommand{\cory}[1]{{\color{red}{\bf cc: #1}}}
\newcommand{\mike}[1]{{\color{blue}{\bf mk: #1}}}

\newcommand{\nilaksh}[1]{{\color{blue}{\bf nd: #1}}}
\newcommand{\mkclean}{
   \renewcommand{\oldinput}[1]{}
   \renewcommand{\hide}[1]{}
   \renewcommand{\todo}[1]{}
   \renewcommand{\old}[1]{}
   \renewcommand{\cory}[1]{}
   \renewcommand{\nilaksh}[1]{}
   \renewcommand{\li}[1]{}
   \renewcommand{\mike}[1]{}
   \renewcommand{\polo}[1]{}
}
\mkclean

\newcommand{\Shield}{\textsc{Shield}}
\newcommand{\Model}{\mathcal{M}}
\newcommand{\origin}{Originative}

\acmDOI{} 
\acmISBN{} 
\acmPrice{}
\acmYear{2019} 
\copyrightyear{2019} 
\setcopyright{rightsretained}
\acmConference[LEMINCS @ KDD'19]{KDD 2019 Workshop on Learning and Mining for Cybersecurity (LEMINCS'19)}{August 5th, 2019}{Anchorage, Alaska, United States}

\begin{document}

\title{The Efficacy of \Shield{} under Different Threat Models}
\subtitle{Paper Type: Appraisal Paper of Existing Method}

\author{Cory Cornelius}
\email{cory.cornelius@intel.com}
\affiliation{
  \institution{Intel Corporation}
  \city{Hillsboro}
  \state{OR}
  \country{USA}
}

\author{Nilaksh Das}
\email{nilakshdas@gatech.edu}
\affiliation{
  \institution{Georgia Institute of Technology}
  \city{Atlanta}
  \state{GA}
  \country{USA}
}

\author{Shang-Tse Chen}
\email{schen351@gatech.edu}
\affiliation{
  \institution{Georgia Institute of Technology}
  \city{Atlanta}
  \state{GA}
  \country{USA}
}

\author{Li Chen}
\email{li.chen@intel.com}
\affiliation{
  \institution{Intel Corporation}
  \city{Hillsboro}
  \state{OR}
  \country{USA}
}

\author{Michael E. Kounavis}
\email{michael.e.kounavis@intel.com}
\affiliation{
  \institution{Intel Corporation}
  \city{Hillsboro}
  \state{OR}
  \country{USA}
}

\author{Duen Horng (Polo) Chau}
\email{polo@gatech.edu}
\affiliation{
  \institution{Georgia Institute of Technology}
  \city{Atlanta}
  \state{GA}
  \country{USA}
}

\renewcommand{\shortauthors}{Cornelius, Das, et al.} 
\renewcommand{\shorttitle}{The Efficacy of SHIELD under Different Threat Models}


\begin{abstract}
\input{000-abstract.tex}
\end{abstract}


\maketitle


\section{Introduction}
\input{100-intro.tex}


\section{Background}

\subsection{\Shield}
\input{210-shield.tex}

\subsubsection{Stochastic Local Quantization (SLQ)}
\input{211-slq.tex}

\subsubsection{Training Models on JPEG-compressed Images}
\input{212-training.tex}

\subsection{Projected Gradient Descent}
\input{220-pgd.tex}


\section{Appraisal Methodology}
\input{300-method.tex}

\subsection{Studying Effect of Training Procedure on Ensemble Robustness}
\input{310-retraining_and_correlation.tex}

\subsection{Extending to More Threat Models}
\input{320-threat_models.tex}

\subsection{Constructing and Evaluating Adaptive Attacks}
\input{330-adaptive.tex}


\section{Results}
\input{400-results.tex}
\subsection{White-box Attack (TM-White)}
\input{410-whitebox.tex}

\subsection{Weak Gray-box attack (TM-Gray1)}
\input{420-gray1.tex}

\subsection{Moderate Gray-box attack (TM-Gray2)}
\input{430-table.tex}
\input{430-gray2.tex}

\subsection{Strong Gray-box attack (TM-\Shield)}
\input{440-shield.tex}


\section{Conclusion}
\input{600-conclusion.tex}


\section*{Acknowledgments}
We thank Richard Shin for sharing his differentiable JPEG implementation.


\bibliographystyle{ieeetr}
\bibliography{bibliography}


\newpage
\appendix

\section{Implementation Issues}
\input{999-100-shield_impl.tex}

\end{document}

%% file: 000-abstract.tex
In this appraisal paper, 
we evaluate the efficacy of \Shield{}~\cite{das2018shield}, 
a compression-based defense framework for countering adversarial attacks on image classification models, which was published at KDD 2018.
Here, we consider alternative threat models not studied in the original work,
where we assume that an \textit{adaptive} adversary is aware of 
the ensemble defense approach, 
the defensive pre-processing, 
and the architecture and weights of the models used in the ensemble.
We define scenarios with varying levels of threat
and empirically analyze the proposed defense 
by varying the degree of information available to the attacker,
spanning from a full white-box attack 
to the gray-box threat model described in the original work.
To evaluate the robustness of the defense against an adaptive attacker,
we consider the \textit{targeted-attack success rate} of the Projected Gradient Descent (PGD) attack,
which is a strong gradient-based adversarial attack proposed in adversarial machine learning research.
We also experiment with training the \Shield{} ensemble from scratch, which is different from re-training using a pre-trained model as done in the original work. 
We find that the targeted PGD attack has a success rate of 64.3\% against the original \Shield{} ensemble in the full white box scenario, but this drops to 48.9\% if the models used in the ensemble are trained from scratch instead of being retrained. Our experiments further reveal that an ensemble whose models are re-trained indeed have higher correlation in the cosine similarity space, and models that are trained from scratch are less vulnerable to targeted attacks in the white-box and gray-box scenarios.

%% file: 100-intro.tex
Adversarial examples are inputs that appear innocuous to humans but fool a machine learning model into making incorrect predictions~\cite{goodfellow2018darkarts}.
Even the most accurate image classification models that rely upon state-of-the-art deep neural networks are potentially vulnerable to these maliciously crafted inputs~\cite{szegedy2014intriguing,madry2018adversarial,carlini2017bypassing}.
Thus, it is an important research problem to develop defenses against such adversarial examples, especially when these models are deployed in safety or security critical systems.

One such recently proposed defense is \Shield{}, 
\textit{Secure Heterogeneous Image Ensemble with Localized Denoising} ~\cite{das2018shield}, 
which uses a combination of techniques to defend against adversarial examples for image classification systems.
\Shield{} trains several image classification models using JPEG-compressed images at different compression levels.
These JPEG-trained models are used in a majority vote ensemble to yield the final classification result for a given input image.
At inference time, \Shield{} also applies \textit{Stochastic Local Quantization} (SLQ), which is a novel randomized form of JPEG compression, as a pre-processing step.
In the face of a non-adaptive adversary that is only aware of the model architecture (but not its weights nor any defensive measures), \Shield{} reports a $16.3\%$ decrease in test accuracy against adversarial examples as compared to a 57.19\% drop when there is no defense.

While experiment results in the original \Shield{} paper emphasized its fast inference and practicality of being a readily available technique, 
it did not consider a full range of threat models or more recent stronger attacks.
Thus, in this work, we aim to appraise \Shield{} and make the following contributions:

\medskip
\noindent \textbf{Evaluating Threat Models with Adaptive Attacker.}
In this work, we further test the robustness of \Shield{}
against an \textit{adaptive} adversary that has access 
to more parts of the defense than originally studied. 
We start with the white-box attack, in which the attacker has access 
to all parts of the inference pipeline, 
and peel back information given to the attacker
by defining varying levels of threat models
that are different from the gray-box threat model analyzed in~\cite{das2018shield}.
We also find that \Shield{} can be made more robust to targeted attacks even in the white-box case by using models in the ensemble whose weights are less correlated.

\medskip
\noindent \textbf{Evaluating Stronger Attacks.}
Since around the time \Shield{} was published, the Projected Gradient Descent (PGD)~\cite{madry2018adversarial} attack had started to gain reputation in becoming one of the strongest gradient-based iterative attacks proposed in adversarial machine learning research.
As PGD was not evaluated in the original \Shield{} paper,
in this work, we contribute evaluation focusing on PGD.
Specifically, we use the \textit{targeted} version of 
PGD to study the robustness of the defense framework under different threat models. 
We focus on reporting the \textit{attack success rate} for targeting the \textit{least-likely} model prediction,
as it is more compelling if an attack is able to affect the prediction of a model significantly, opposed to only slightly changing it.
Since we target the least likely model prediction for the attack, 
it is only easier for the adversary to target other predictions as they already have a higher rank in the prediction likelihood.

\medskip
\noindent \textbf{Analyzing the Effect of Training Procedure on Robustness of \Shield{}.}
We take a closer look into the training procedure of \Shield{}, whose original approach of creating an ensemble of JPEG-trained models 
involves re-training the models from a pre-trained ResNet-50 v2 model.
We find that this re-training approach yields an ensemble of models whose weights are highly correlated,
making the defense less robust to attacks
when the attacker has access to one or more of the models in the ensemble.
To explore how to alleviate this issue, 
we experiment with newly trained models 
that have been trained on JPEG-compressed images from scratch,
and perform the same analysis for comparison.
We find that the training procedure indeed affects the robustness of the defense at reducing the success of strong targeted attacks. In the full white-box case, simply training the models of the ensemble from scratch can reduce the attack success rate by 24\%.

%% file: 210-shield.tex
\Shield{} is a defense framework published in KDD 2018 that employs (1) Stochastic Local Quantization and central cropping 
as pre-processing; (2) model re-training; and (3) an ensemble of models to defend against adversarial attacks on image classification systems. Below, we summarize the various components of the \Shield{} defense framework.

%% file: 211-slq.tex
SLQ is a randomized form of image compression introduced in \cite{das2018shield}. This method leverages JPEG compression as a form of defense. Since JPEG compression performs quantization in the frequency domain, the authors hypothesize that this step also removes adversarial perturbations from the image. The authors also argue that since the quantization step is non-differentiable, the attacker cannot obtain any useful gradients to directly perform gradient-based attacks. 
Additionally, SLQ uses randomization to further mask the gradients by breaking up the image into smaller blocks, and applying a different JPEG compression level to each block.

More concretely, 
let us denote the input image as $x$, 
and the JPEG compression at quality level $q$ 
applied to the input image as $JPEG(x, q)$. 
SLQ uses $K$ qualities $\{q_1, \dots, q_K\}$ 
to compute $JPEG(x, q_1)$, $\dots$, $ JPEG(x, q_K)$,
and use these images to randomly pick patches
from the corresponding locations to stitch up the final image.

%% file: 212-training.tex
Since \Shield{} employs SLQ pre-processing that uses different levels of JPEG compression, the authors of \cite{das2018shield} also re-train the image classification model to be more robust to the pre-processing operation. Specifically, \Shield{} takes a pre-trained ResNet-50 v2 model, denoted as $\Model$, and re-trains it on images compressed at JPEG quality $q$ to obtain a new ResNet-50 v2 model $\Model_q$. The weights of $\Model$ are used as the initial weights while training $\Model_q$ to get faster convergence.
\Shield{} trains multiple models in this manner using different JPEG quality levels $\{q_1, \dots, q_K\}$ to obtain $\Model_{q_1}, \dots, \Model_{q_K}$, which are then used as a majority vote ensemble.

%% file: 220-pgd.tex
Given a benign input instance $x$, a targeted adversarial attack aims to find a small perturbation $\delta$ that changes the prediction of model $\Model$ to a target class $t$ different from the true class $y$, i.e., $\Model(x+\delta) = t$, where $t\neq y, ||\delta||\le\epsilon$. We call $\epsilon$ the perturbation strength with respect to a specific norm $||\cdot||$. 
In this work we only examine strength in terms of $\ell_\infty$ because it is easy to understand---the maximum pixel value deviation an attacker can apply across all image channels.

\textit{Projected Gradient Descent} (PGD)~\cite{madry2018adversarial} is one of the strongest first-order attacks in the adversarial machine learning literature.
PGD iteratively minimizes a loss function $L(x, t)$. In this work, we use 
cross-entropy for $L$ that computes the distance between the softmax of the logit layer and the one-hot representation of the target class $t$.
We denote the logit layer of model $\Model$ as $f(x, \Model)$.
 In each iteration $i$, it computes the direction of perturbation by taking the sign of the gradient of loss
function $L$ with respect to the current perturbed instance $x^i$, and then perform a projection step back to the feasible set, i.e., within $\epsilon$ ball of the original instance while remaining a valid image:
\[
x^{i+1} = Proj\left( x^i - \alpha \cdot \text{sign} (\nabla_{x^i} L(x^i, t) ) \right)
\]
Instead of starting from the original instance, we use a random starting point within $\epsilon$ ball of the original instance~\cite{madry2018adversarial}.
Following~\cite{kurakin2016adversarial}, we use the least likely class $y_{LL}$ predicted by the model as our target class.

%% file: 300-method.tex
When deploying a defense to mitigate adversarial examples, one must understand the limits of that defense in terms of the strength of the attacker and threat models it was tested for.
We usually quantify a defense's capability as a well-defined threat model along with a security curve that plots the accuracy of the model in the face of an adversary that has the ability to manipulate the image at different strengths for that threat model~\cite{biggio2017patterns}.
We measure strength as the difference between a test image and its corresponding adversarial image generated via an attack method.
Such differences are high-dimensional so we often summarize this difference by computing some metric like $\ell_\infty$ (largest absolute difference) or $\ell_2$ (euclidean distance).
At the limit of the strength (i.e., the ability to manipulate pixels with any value), the attacker can simply use those examples that are already incorrectly classified or just replace the image with another image of their desired target.
It turns out, however, that many models exhibit weaknesses at much smaller strengths where it is difficult for humans to visually distinguish any adversarial perturbation.
Note that the study of these security curves is of interest to defenders due to the asymmetry between adversaries and defenders.
That is, adversaries need only find one particular weakness while defenders ought to mitigate all possible weaknesses.

\begin{figure}[t]
\centering
\begin{subfigure}[b]{\linewidth}
  \includegraphics[width=1\linewidth,trim={0.5cm 0 0 0},clip]{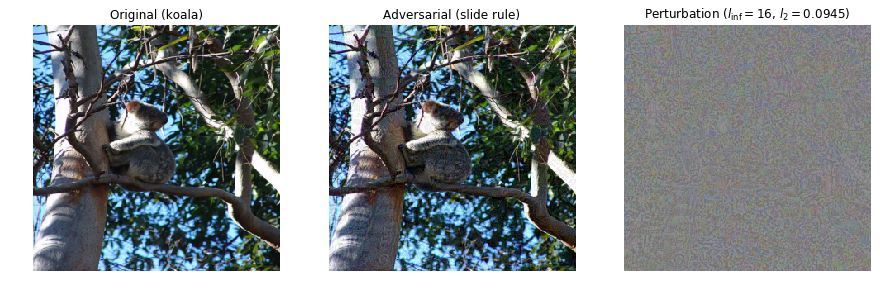}
  \caption{}
  \label{fig:koala} 
\end{subfigure}

\begin{subfigure}[b]{\linewidth}
  \includegraphics[width=1\linewidth,trim={0.5cm 0 0 0},clip]{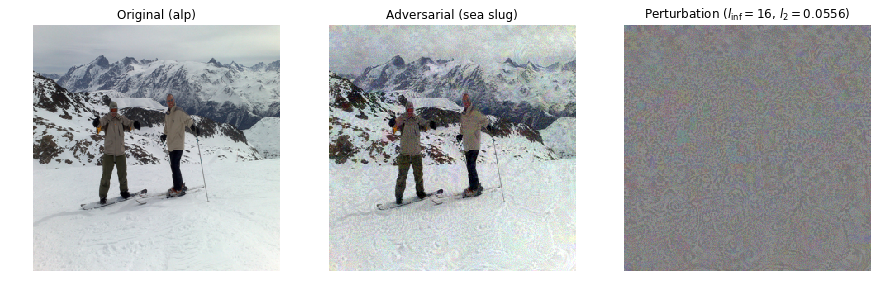}
  \caption{}
  \label{fig:alp}
\end{subfigure}

\caption{Example adversarial images generated against \Shield{}. The first column shows the original images and their corresponding predictions. 
The second column shows the adversarial images their corresponding predictions. The final image is the difference between the original and adversarial image, which we call the perturbation. Note that both perturbations have $\ell_\infty$ distance of 16 but different $\ell_2$ distances. The perturbation in the \textit{alp} photo is visually noticeable, while the perturbation in the \textit{koala} photo is not. Despite being visually noticeable, the \textit{alp} perturbation has lower $\ell_2$ than the \textit{koala} perturbation. Ideally a chosen distance would be well ordered with respect to visual distinguishability. Images are best viewed in color.}
\label{fig:images}
\end{figure}

In this work, we examine strength in terms of $\ell_\infty$.
Figure~\ref{fig:images} shows two different adversarial images with their corresponding perturbations.
Both perturbations have $\ell_\infty$ of $16$ yet exhibit different $\ell_2$ distances.
The $\ell_2$ distance of the \textit{koala} perturbation is larger than the $\ell_2$ distance of the \textit{alps} perturbation, yet the \textit{alps} perturbation is easily discernible in the adversarial image.
Finding better metrics to summarize adversarial perturbations that take into account human perception remains an open problem.
Because \Shield{} was evaluated against ImageNet, we chose the same attacker strength that \textit{other} state-of-the-art defenses on ImageNet use~\cite{Kurakin2017adv}: an $\ell_\infty$ of $16$ out of $255$.

%% file: 310-retraining_and_correlation.tex
Since the \Shield{} framework deploys an ensemble of models re-trained from the same ResNet-50 v2 model $\Model$, we posit that the model weights may be highly correlated, making it easier for the attacker to attack the ensemble even under weak gray-box restrictions in which the attacker has access to only some of the models from the ensemble. 
To study the effect of this potential issue, we train the models from scratch on JPEG-compressed images of corresponding qualities used in the original work.

In \cite{das2018shield}, the model \textit{re-trained} from a pre-trained ResNet-50 v2 model $\Model$ using images with JPEG quality $q$ were denoted as $\Model_q$. We call these models as ``\textbf{Derivative}'' models.
Adopting a similar notation,
we call the new ResNet-50 v2 model trained from scratch 
on images with JPEG quality $q$ as $\Model^*_q$,
and we refer to them as ``\textbf{\origin}'' models.
Hence, we perform our analysis on two ensembles, the newly trained $\{\Model^*_{20}$, $\Model^*_{40}$, $\Model^*_{60}$, $\Model^*_{80}\}$ (or \Shield{}-\origin) and the original $\{\Model_{20}$, $\Model_{40}$, $\Model_{60}$, $\Model_{80}\}$ (or \Shield{}-Derivative).

Also, to quantify the correlation between 2 models, we flatten, normalize and concatenate the weights from each layer of a model to form a single vector, and calculate the cosine similarity with the corresponding vector for the other model. Indeed, we find that \Shield{}-Derivative (which has been obtained using initial weights from the same pre-trained ResNet-50 v2 model) has a higher average cosine similarity of 0.64 between any pair of models in the ensemble. Whereas, \Shield{}-\origin{} (whose models have been trained from scratch) has a lower average cosine similarity of 0.42 between the pairs of models. One can further decrease this correlation by explicitly adding the similarity measure to the loss function as a regularizer, and train an ensemble jointly to make it more robust to transferability attacks.

%% file: 320-threat_models.tex
\begin{table*}[htp]
\centering

\begin{tabular}{lcccc}
\toprule
             & \multicolumn{4}{c}{Attacker has access to}                                           \\
\cmidrule(lr){2-5}
Threat Model (TM) & Model Architecture & Weights of Models in Ensemble           & Defense Strategy & Defense Parameters \\
\midrule
TM-White     & Yes                & All          & Yes              & Yes                \\
TM-Gray1     & Yes                & Some Models & Yes              & Yes                \\
TM-Gray2     & Yes                & No  & Yes              & Yes                \\
TM-\Shield    & Yes                & No  & No               & No                    \\
\bottomrule
\end{tabular}

\caption{Threat models studied in this work, ordered by decreasing severity. We define different scenarios with varying levels of threat. TM-White (top) allows the attacker to access the whole inference pipeline including the defense parameters, whereas TM-\Shield{} (bottom), originally studied in \cite{das2018shield}, only provides the model architecture to the attacker.}
\label{tab:threat}

\end{table*}

To fully understand the robustness of the proposed defense under different circumstances, it is important to set meaningful restrictions on the capabilities of the attacker according to the application domain \cite{gilmer2018motivating}.
Hence, we define varying degrees of threat models, and evaluate the proposed \Shield{} defense under different scenarios. 
These threat models vary from a full white-box attack (no restrictions) to strong gray-box restrictions as originally studied in \cite{das2018shield}.
Table~\ref{tab:threat} summarizes these threat models, and we describe them in more detail below.

\smallskip
\noindent \textbf{TM-White}.
This threat model corresponds to the full white-box attack that imposes no restrictions on the attacker, who is allowed to access to all parts of the proposed defense. For \Shield{}, this includes the model architecture, weights, any JPEG pre-processing (i.e., SLQ), as well as all the defense parameters (e.g., SLQ in \cite{das2018shield} uses JPEG qualities 20, 40, 60 and 80).
In practice, attackers typically do not have full access to all information of the defense (if they do, that could mean an attacker is in complete control of the defender and could do anything they want). 
However, studying white-box attacks help researchers better understand a defense's robustness under the most hostile situation.

\medskip
\noindent \textbf{TM-Gray1}.
This is a gray-box threat model that imposes weak restrictions on the attacker. Namely, the attacker now only has access to \textit{some} of the models' weights from the ensemble. The attacker is still aware of the defense strategy and the defense parameters. Since \Shield{} uses 4 models in the ensemble, we study the robustness of the defense by varying the number of models that the attacker can access, and report the average attack success rate. For example, if the attacker can access 2 out of the 4 models from the ensemble, this yields 6 combinations of models that the attacker can attack,
i.e., ${4 \choose 2} = 6$, 
and we report the average success rate from all the 6 attack trials in this case.

\medskip
\noindent \textbf{TM-Gray2}.
This is a moderate gray-box threat model that imposes some more restrictions on the attacker. Specifically, now the attacker is aware of the the model architecture and the defense strategy and parameters, but cannot access any of the model weights and is unaware of the training strategy. In this scenario, the attacker is faced with training their own models on JPEG-compressed images. 
We evaluate this scenario by creating adversarial examples using the newly trained \textit{\origin{}} models $\{\Model^*_q\}$ and testing them against with \Shield{}'s default \textit{Derivative} models $\{\Model_q\}$.

\medskip
\noindent \textbf{TM-\Shield{}}.
This corresponds to the original gray-box threat model studied in \cite{das2018shield} which imposes strong restrictions on the attacker. In this scenario, the attacker is only aware of the model architecture used in the ensemble and is oblivious of the defense pre-processing and does not have access to the model weights. 
\\

%% file: 330-adaptive.tex
\Shield{} integrates several techniques that mask the gradients of a model so that an adversary cannot perform an exact gradient-based attack that uses backpropagation. 
This backpropagation computation is exactly the same method one would use to learn such a model, except that rather than changing the weights of the model, the attacker seeks to change only the input image.
For this work, we rely upon the PGD attack as implemented in Cleverhans~\cite{papernot2018cleverhans} (which contains technical details).

Backpropagation requires that all operations in the model be differentiable.
By using JPEG compression, which is non-differentiable, 
\Shield{} forces attackers to create JPEG compression approximations that are differentiable.
A recent work~\cite{shin2018jpeg} describes how to do this, 
and we apply the authors' differentiable JPEG approximation to attack \Shield{}.
Similarly, because the majority voting of the ensemble of models in \Shield{} is also non-differentiable, 
we approximate the majority vote by averaging the logits output of each model before applying the softmax function.
One can also apply alternative ensemble approximations~\cite{he2017ensemble}, 
but we found averaging to be effective.

A further difficulty in naively applying PGD to \Shield{} is the stochasticity of the SLQ pre-processing step.
Because SLQ introduces randomness in the input presented to the model, the model effectively sees a different, albeit similar, input image every time for the same image.
\Shield{} relies upon its ensemble of JPEG trained models to ensure that the same input image even after SLQ pre-processing still maps to the same prediction.
The attacker, however, typically has no control of this randomness.
To circumvent this issue, 
we compress the image at multiple compression levels --- $20$, $40$, $60$ and $80$ (the JPEG qualities used by SLQ) --- and average over these compression levels to find an adversarial perturbation.
More concretely, for the purpose of attacking, we compute the logits of a given ensemble $\mathcal{E}$ for the input $x$ as,
$$
\frac{1}{|\mathcal{Q}|\cdot|\mathcal{E}|} \sum_{\Model \in \mathcal{E}} \sum_{q \in \mathcal{Q}} f(JPEG(x, q), \Model)
$$
where $\mathcal{Q}$ is the set of JPEG compression levels used by SLQ and $f(x, \Model)$ is the logits output of the model $\Model$ for input $x$.

%% file: 400-results.tex
For our experiments, we sampled $1000$ of the $50000$ ImageNet \cite{deng2009imagenet} validation images and used PGD with 20 iterations to generate adversarial images. 
We performed the targeted version of the PGD attack with the \textit{least-likely} prediction of the model under attack as the targets. 
We report the \textit{attack success rate}, which is the fraction of images that were incorrectly classified by the model with the target label specifically chosen by the attacker. 
The attack is not successful if the model misclassifies the image but the output label is not the same as the target label. 
In the following section, we now present our results on the threat models being studied.

%% file: 410-whitebox.tex
The results from a full white-box targeted attack are shown in Table~\ref{tab:white}.
We see that only $17$ of the $1000$ adversarially crafted images are correctly classified by the original \Shield{} model (\Shield-Derivative).
The remaining $983$ were incorrectly classified as something other than the ground truth label.

Furthermore, an attacker is able to force \Shield{}-Derivative to make a prediction of their choosing.
Here we chose the prediction target to be the least-likely class that the differentiable version of ensemble predicts on the original image.
Such a target should be difficult for an attacker to hit, yet our results show that $643$ of those $1000$ adversarial image were successfully predicted as the least-likely label. However, we see in the case of \Shield-\origin{} that an ensemble having uncorrelated models can alleviate this problem and increase the defense robustness to a targeted attack. The attacker is only able to target $489$ images out of $1000$ in this case.

\begin{table}[t]
\centering

\begin{tabular}{lrr}
\toprule
Ensemble & White-box Attack Success & Acc. \\
\midrule
\Shield-Derivative & 0.643 & 0.017 \\
\Shield-\origin & \textbf{0.489} & 0.022 \\
\bottomrule
\end{tabular}

\caption{White-box robustness to targeted attacks.  \Shield-\origin{} thwarts more targeted attacks (i.e., attacker is less successful), 
even when adversary has access to the entire inference pipeline, 
since models trained from scratch have lower correlation than models re-trained from the another model as in \Shield-Derivative.
Accuracy of both models are adversely affected.
}
\label{tab:white}

\end{table}

%% file: 420-gray1.tex
\begin{figure}[t]
    \centering
    \includegraphics[width=\linewidth]{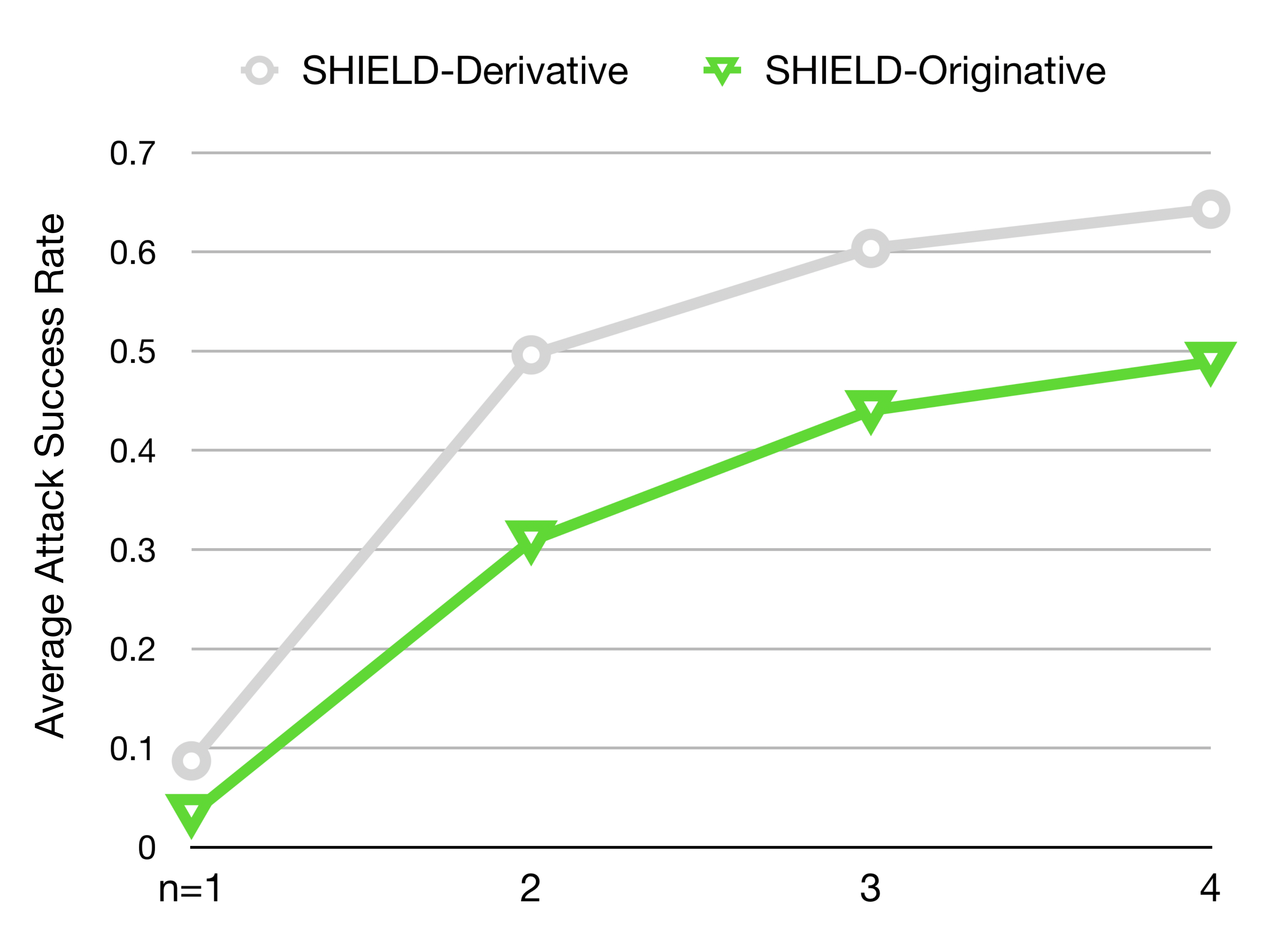}
    \caption{Comparison of the average attack success for TM-Gray1, where the attacker has access to $n$ out 4 models from the ensemble. The attack is least successful in both cases when the attacker has access to only one model, and most successful when the attacker has access to all 4 models (TM-White). Models trained from scratch (\Shield-\origin, in green) are significantly more robust to an adaptive targeted attack as compared to the original \Shield{} ensemble, in gray.}
    \label{fig:gray1}
\end{figure}

In this threat model, the attacker can only access some of the models from the ensemble, while having full knowledge of the defense strategy. We perform this ablation study by averaging over the attack success rate from attacking all possible model combinations corresponding to the attacker's budget constraint (i.e., the number of models the attacker can access). The most restrictive case is when the attacker can only access one of the models. In this scenario, the attacker has 4 choices for performing the attack (on any one of the 4 models from the ensemble). Hence the average attack success is reported from the 4 trials that the attacker can make. Similarly, the attacker has 6 choices when allowed to attack 2 models together, and so on. 

The results of this experiment are shown in Figure~\ref{fig:gray1}, which plots the average attack success from attacking $n$ out of 4 models, by varying $n$ on the x-axis. As is evident, this study spotlights the effect of training from scratch and the risk of having models that are correlated. Note here that TM-White is an extreme case of this threat model when the attacker can access the whole ensemble ($n=4$). 

Even with just one model available to the attacker, the accuracy of the original ensemble (\Shield-Derivative) falls to 20.33\%, and further to 1.7\% when all models are available to the attacker (TM-White). The \Shield-\origin{} ensemble also shows similar trends with respect to accuracy. However, there is a clear contrast when it comes to the attack success rates for the two ensembles. The targeted attack has a lower success rate with the \Shield-\origin{} ensemble, since the weights of the models in this ensemble are less correlated, and hence the attack does not transfer so well to the other models.

%% file: 430-table.tex
\begin{table*}[]
\begin{tabular}{lrrrrrr}
\toprule
 & \multicolumn{1}{l}{} & \multicolumn{5}{c}{Defending Model} \\
\cmidrule(lr){3-7}
Attacked Ensemble & $\#$model known & \Shield-Derivative & $\Model_{20}$ & $\Model_{40}$ & $\Model_{60}$ & $\Model_{80}$ \\
\cmidrule(lr){1-2}\cmidrule(lr){3-7}
\multirow{4}{*}{\Shield-Originative} & 1 &  0.000 (0.303) & 0.000 (0.215) & 0.000 (0.273) & 0.000 (0.272) & 0.000 (0.269)  \\
& 2 &  0.000 (0.265) & 0.000 (0.184) & 0.000 (0.240) & 0.000 (0.237) & 0.000 (0.228)  \\
& 3 &  0.000 (0.231) & 0.000 (0.165) & 0.000 (0.206) & 0.000 (0.211) & 0.000 (0.202)  \\
& 4 &  0.000 (0.214) & 0.000 (0.139) & 0.000 (0.185) & 0.000 (0.198) & 0.000 (0.185)  \\
 \cmidrule(lr){1-2}\cmidrule(lr){3-7}
 \textit{No attack} & & \textit{0.633} & \textit{0.538} & \textit{0.597} & \textit{0.574} & \textit{0.603} \\
\bottomrule
\end{tabular}
\caption{
Average attack success rate, 
and average model accuracy (in parenthesis) 
for the TM-Gray2 threat model. 
The targeted attack does not have any success in this scenario since the attacker does not have access to the model weights and the training procedure, and is thus forced to train their own models which are not guaranteed to transfer even with an adaptive targeted attack.}
\label{tab:gray2}
\end{table*}

%% file: 430-gray2.tex
This is a gray-box threat model that places more restrictions on the attacker. The attack can no longer access the weights of the ensemble and is unaware of the training procedure, but is aware of the model architecture and defense parameters.
We believe such a threat model can be realistic since 
many applications employ existing architectures but with weights learned for their specific task.
Furthermore, weights of a model can remain confidential by using a trusted execution environment or deployment as a cloud API.
Finally, we believe our assumption that the attacker is aware of a JPEG-based defense is reasonable.
Like model architectures, the set of known defenses is small enough that an attacker can exhaustively try them until they are successful.

In this threat model, the attacker is forced to train their own proxy models for performing the attack.
To mimic this, we first generated adversarial examples against newly trained models from the \Shield-\origin{} ensemble using the adaptive attack procedure described above.
We then fed these adversarial examples into the original \Shield-Derivative model to determine whether they fool \Shield{}. 
We also evaluate these attacked images using the original JPEG-trained models ($\Model_{20}$, $\Model_{40}$, $\Model_{60}$, $\Model_{80}$) for comparison.

The results from this experiment are shown in Table~\ref{tab:gray2}. In all of these tests, we found that the adversary was unable to control the prediction target.
That is, the adversarial image we generated did not classify to the ground truth label nor the targeted label, but to some other label.
It is application dependent whether this ability of the attacker to control the predicted output is of concern.
In authentication usages, for example, the attacker might impersonate a specific user with targeted attacks or simply any authorized user with untargeted attacks. 

We also tested the Fast Gradient Method (FGM) \cite{goodfellow2014explaining} which is known to be  transferable even across model architectures. Our results showed that FGM targeted attack in this scenario is still not successful in forcing a prediction on the ensemble, resulting in a $0\%$ attack success rate, although it affects the model accuracy more adversely than PGD.
 
 In this gray-box threat model, we see a degradation in the model accuracy of the ensemble as well as the JPEG-trained models; however, an attacker cannot control the prediction like they could in the white-box threat model.

%% file: 440-shield.tex
This is the threat model originally studied in \cite{das2018shield} that places stronger gray-box restrictions on the attacker. The attacker in this scenario is only aware of the model architecture and cannot access any other parts of the inference pipeline. This threat model assumes that the attacker is non-adaptive. 
To mimic this, we turn off the differentiable JPEG component in our adaptive attack computation and attack a ResNet-50 v2 model that is not part of the original \Shield{} ensemble.
Although we observe a drop in model accuracy from 0.633 to 0.381 for the \Shield-Derivative ensemble and from 0.77 to 0.423 for the \Shield-\origin{} ensemble due to the added perturbation,
we find that the non-adaptive targeted PGD attack is not successful with a $0\%$ attack success.

%% file: 600-conclusion.tex
As with any empirical security analysis, our results represent upper-bounds on the robustness of \Shield{}.
Attacks only get stronger.
We believe that an adversary can better target gray-box attacks using a technique that recently won the both the non-targeted adversarial attack and targeted adversarial attack competitions at the 2017 Competition on Adversarial Attacks and Defenses~\cite{dong2018boosting}, or adversaries will use query-based black-box methods to increase their targeted attack success~\cite{ilyas2018blackbox}.
More so, while understanding the limits of a model with respect to some $\ell_\infty$ limit is instructive, ultimately adversaries will find more clever ways to fool a model~\cite{sharma2018attacking,brown2017adversarial}.

Our main goal of this work is to understand the efficacy of \Shield{} against adaptive attack techniques and under different threat models. We evaluate the robustness of the ensemble-based defense at resisting against targeted attacks. 
Our results show that \Shield{} can be made more robust to such attacks in the white-box scenario by modifying the training procedure so as to deploy models whose weights are less correlated. This is further supported by our gray-box experiments, wherein the models trained from scratch are more robust to making maliciously targeted predictions. In this work we simply train the models from scratch on different data distributions (by varying JPEG compression levels) to obtain models that have less correlated weights, but one can imagine a training paradigm that explicitly models this as a regularizer in the training optimization criteria.
We hope that future defenses will employ similar adaptive attack techniques to demonstrate their robustness at a variety of attack strengths and in the face of different, perhaps more realistic, threat models.

%% file: 999-100-shield_impl.tex
The first author of this work discovered an issue 
in the original \Shield{} evaluation code implementation that affects the results reported in the published paper~\cite{das2018shield}.
During training, central crops are enabled to better learn discriminative features for the object of interest.
However, during evaluation it is customary to turn off this central cropping.
The implementation of \Shield{} does not turn off this cropping during evaluation nor did the attacks in \Shield{} take into account this cropping when generating perturbations.

The first author of this work discovered this issue after noticing adversarial images were not working against the public implementation of \Shield{}.
He found that the images were adversarial against another implementation of \Shield{} using the same SLQ and model parameters, but not against the originally released \Shield{} implementation\footnote{\url{https://github.com/poloclub/jpeg-defense} \\ at commit \texttt{1576429cf199c38065b941a48b0fcd7747901457}}.
After some investigation, he found the aforementioned central-cropping-at-evaluation-time issue, disabled this central cropping, and found that the adversarial images remained adversarial.

As such, central cropping is now a feature of \Shield{} in the sense that the evaluation portion of the published paper~\cite{das2018shield} is SLQ \textit{with central cropping}.
Further experiments reveal that central cropping contributes significantly to the reported robustness of \Shield{}. 
In this work,
we do not include this cropping as part of the pre-processing step, and only consider SLQ as is.

%% file: main.bbl
\begin{thebibliography}{10}

\bibitem{das2018shield}
N.~Das, M.~Shanbhogue, S.-T. Chen, F.~Hohman, S.~Li, L.~Chen, M.~E. Kounavis,
  and D.~H. Chau, ``Shield: Fast, practical defense and vaccination for deep
  learning using jpeg compression,'' in {\em Proceedings of the 24th ACM SIGKDD
  International Conference on Knowledge Discovery \& Data Mining ({KDD})}, Aug.
  2018.
\newblock \url{http://doi.acm.org/10.1145/3219819.3219910}.

\bibitem{goodfellow2018darkarts}
I.~J. Goodfellow, ``Defense against the dark arts: An overview of adversarial
  example security research and future research directions,'' {\em CoRR},
  vol.~abs/1806.04169, 2018.
\newblock \url{https://arxiv.org/abs/1806.04169}.

\bibitem{szegedy2014intriguing}
C.~Szegedy, W.~Zaremba, I.~Sutskever, J.~Bruna, D.~Erhan, I.~J. Goodfellow, and
  R.~Fergus, ``Intriguing properties of neural networks,'' in {\em Proceedings
  of the 2nd International Conference on Learning Representations ({ICLR})},
  2014.
\newblock \url{https://openreview.net/forum?id=kklr_MTHMRQjG}.

\bibitem{madry2018adversarial}
A.~Madry, A.~Makelov, L.~Schmidt, D.~Tsipras, and A.~Vladu, ``Towards deep
  learning models resistant to adversarial attacks,'' in {\em Proceedings of
  the 6th International Conference on Learning Representations ({ICLR})}, Apr.
  2018.
\newblock \url{https://openreview.net/forum?id=rJzIBfZAb}.

\bibitem{carlini2017bypassing}
N.~Carlini and D.~Wagner, ``Adversarial examples are not easily detected:
  Bypassing ten detection methods,'' in {\em Proceedings of the 10th ACM
  Workshop on Artificial Intelligence and Security ({AISec})}, Nov. 2017.
\newblock \url{http://doi.acm.org/10.1145/3128572.3140444}.

\bibitem{kurakin2016adversarial}
A.~Kurakin, I.~Goodfellow, and S.~Bengio, ``Adversarial examples in the
  physical world,'' {\em arXiv preprint arXiv:1607.02533}, 2016.

\bibitem{biggio2017patterns}
B.~Biggio and F.~Roli, ``Wild patterns: Ten years after the rise of adversarial
  machine learning,'' {\em CoRR}, vol.~abs/1712.03141, 2017.
\newblock \url{https://arxiv.org/abs/1712.03141}.

\bibitem{Kurakin2017adv}
A.~Kurakin, I.~J. Goodfellow, and S.~Bengio, ``Adversarial machine learning at
  scale,'' in {\em International Conference on Learning Representations}, 2017.

\bibitem{gilmer2018motivating}
J.~Gilmer, R.~P. Adams, I.~Goodfellow, D.~Andersen, and G.~E. Dahl,
  ``Motivating the rules of the game for adversarial example research,'' {\em
  arXiv preprint arXiv:1807.06732}, 2018.

\bibitem{papernot2018cleverhans}
N.~Papernot, F.~Faghri, N.~Carlini, I.~Goodfellow, R.~Feinman, A.~Kurakin,
  C.~Xie, Y.~Sharma, T.~Brown, A.~Roy, A.~Matyasko, V.~Behzadan,
  K.~Hambardzumyan, Z.~Zhang, Y.-L. Juang, Z.~Li, R.~Sheatsley, A.~Garg,
  J.~Uesato, W.~Gierke, Y.~Dong, D.~Berthelot, P.~Hendricks, J.~Rauber, and
  R.~Long, ``Technical report on the cleverhans v2.1.0 adversarial examples
  library,'' {\em arXiv preprint arXiv:1610.00768}, 2018.
\newblock \url{http://arxiv.org/abs/1610.00768}.

\bibitem{shin2018jpeg}
R.~Shin and D.~Song, ``Jpeg-resistant adversarial images,'' in {\em Proceedings
  of the Machine Learning and Computer Security Workshop}, Dec. 2017.
\newblock
  \url{https://machine-learning-and-security.github.io/papers/mlsec17_paper_54.pdf}.

\bibitem{he2017ensemble}
W.~He, J.~Wei, X.~Chen, N.~Carlini, and D.~Song, ``Adversarial example defense:
  Ensembles of weak defenses are not strong,'' in {\em Proceedings of the 11th
  {USENIX} Workshop on Offensive Technologies ({WOOT})}, (Vancouver, BC),
  {USENIX} Association, 2017.
\newblock
  \url{https://www.usenix.org/conference/woot17/workshop-program/presentation/he}.

\bibitem{deng2009imagenet}
J.~Deng, W.~Dong, R.~Socher, L.-J. Li, K.~Li, and L.~Fei-Fei, ``Imagenet: A
  large-scale hierarchical image database,'' in {\em IEEE Conference on
  Computer Vision and Pattern Recognition}, pp.~248--255, 2009.

\bibitem{goodfellow2014explaining}
I.~J. Goodfellow, J.~Shlens, and C.~Szegedy, ``Explaining and harnessing
  adversarial examples,'' {\em arXiv preprint arXiv:1412.6572}, 2014.

\bibitem{dong2018boosting}
Y.~Dong, F.~Liao, T.~Pang, H.~Su, J.~Zhu, X.~Hu, and J.~Li, ``Boosting
  adversarial attacks with momentum,'' in {\em Proceedings of the 2018 IEEE/CVF
  Conference on Computer Vision and Pattern Recognition ({CVPR})}, 2018.
\newblock \url{https://doi.org/10.1109/CVPR.2018.00957}.

\bibitem{ilyas2018blackbox}
A.~Ilyas, L.~Engstrom, A.~Athalye, and J.~Lin, ``Black-box adversarial attacks
  with limited queries and information,'' in {\em Proceedings of the 35th
  International Conference on Machine Learning ({ICML})}, July 2018.
\newblock \url{http://proceedings.mlr.press/v80/ilyas18a.html}.

\bibitem{sharma2018attacking}
Y.~Sharma and P.-Y. Chen, ``Attacking the madry defense model with $l_1$-based
  adversarial examples,'' in {\em Proceedings of the 6th International
  Conference on Learning Representations ({ICLR})}, Apr. 2018.
\newblock \url{https://openreview.net/forum?id=Sy8WeUJPf}.

\bibitem{brown2017adversarial}
T.~B. Brown, D.~Man{\'e}, A.~Roy, M.~Abadi, and J.~Gilmer, ``Adversarial
  patch,'' in {\em Proceedings of the Machine Learning and Computer Security
  Workshop}, Dec. 2017.
\newblock
  \url{https://machine-learning-and-security.github.io/papers/mlsec17_paper_27.pdf}.

\end{thebibliography}
